\documentclass[review]{elsarticle}

\usepackage{lineno,hyperref}
\usepackage{multirow}

\usepackage{times}
\usepackage{latexsym}
\usepackage{graphicx}
\usepackage{caption}
\usepackage{subcaption}
\usepackage{algorithm}
\usepackage{algorithmic}
\usepackage{amsmath}

\graphicspath{{./imgs/}}
\usepackage{booktabs}
\usepackage{tabularx}

\newcommand\clearrow{\global\let\rowmac\relax}
\clearrow

\journal{Journal of Cognitive Processing}

\begin{document}

\begin{frontmatter}

\title{Online recognition of unsegmented actions with hierarchical SOM architecture}

\author{Zahra Gharaee\corref{mycorrespondingauthor}}
\address{Cognitive Science, Department of Philosophy \\ University of Lund, Lund, Sweden}\fnref{mycorrespondingauthor}
\address{Computer Vision Labarotory (CVL), Department of Electrical
	Engineering \\ University of Link\"{o}ping, Link\"{o}ping, Sweden }\fnref{mycorrespondingauthor}
\fntext[myfootnote]{Computer Vision Labarotory (CVL)\\ Link\"{o}ping University
	58183 Link\"{o}ping, Sweden}


\cortext[mycorrespondingauthor]{Corresponding author}
\ead{zahra.gharaee@liu.se}


\begin{abstract}
Automatic recognition of an online series of unsegmented actions requires a method for segmentation that determines when an action starts and when it ends. In this paper, a novel approach for recognizing unsegmented actions in online test experiments is proposed. The method uses self-organizing neural networks to build a three-layer cognitive architecture.  The unique features of an action sequence are represented as a series of elicited key activations by the first-layer self-organizing map. An average length of a key activation vector is calculated for all action sequences in a training set and adjusted in learning trials to generate input patterns to the second-layer self-organizing map. The pattern vectors are clustered in the second layer and the clusters are then labeled by an action identity in the third layer neural network. The experiment results showing that although the performance drop slightly in online experiments compared to the offline tests, the ability of the proposed architecture to deal with the unsegmented action sequences as well as the online performance make the system more plausible and practical in real case scenarios.

\end{abstract}


\end{frontmatter}


\section{Introduction}
\label{intro}
Action recognition plays an important role in interaction between any two agents. Human-robot interactions requires that the robot can recognize what kind of action the human is performing. There are several other applications for action recognition systems including human-computer interaction, video retrieval, sign language recognition, medical health care, video analysis (sports video analysis), game industry and video surveillance. In earlier works \cite{Gharaee2,Gharaee3,Gharaee4,Gharaee5,Gharaee6}, Gharaee et al., developed a system for human action recognition using hierarchical architecture based on self-organizing neural networks.

As a background to the method proposed in this article there is a presentation of some psychological approaches to action categorization and to event segmentation. There is also a description of a few earlier computational attempts to solve the action segmentation problem. The rest of the paper is organized as follows: The proposed architecture is described in section 2, the experiments on action recognition are presented in detail in section 3, a discussion about the method proposed of this paper with comparison to some other techniques is presented in section 4 and finally section 5 concludes the paper.

\subsection{Psychological approaches of event perception and segmentation}
Michotte \cite{Michotte} conducted a series of early studies concerning interactions between two objects and he argued that a causal interaction between two objects is perceived when we see the motion of the objects as a single event. Another early contributor was Gibson \cite{Gibson1,Gibson2}. He identified three kinds of events in visual perception: (1) changes in the layout of the surfaces, (2) changes in the color or texture of the surfaces and (3) the coming into or out of existence of surfaces. He argued that the presence of an invariant structure persisting throughout the change is the main factor in creating an event.

A third approach to event perception, based on biological motion, originates from the studies by \cite{johansson}. He developed a method called the patch-light technique in which reflective patches are placed on the body of a subject that performs different actions. The subject is filmed in high-contrast light condition and the film is shown to observers. The observers could only see the movements of the patch-light points, but they could nevertheless recognize, within tenths of a second, that the moving light points come from a human performing an action such as walking or crawling. \cite{Gardenfors1,Gardenfors2} generalize Johansson’s approach, proposing that human cognition represents an action by the pattern of forces generating it. 

A common feature of these three approaches to event perception is that the dynamic features of the activity are critical for perceiving and categorizing events. At the same time they indicate that there is a higher-order stability in events that persist through these changes.

The problem of human event segmentation concerns how our perceptual system can partition the stream of experience into meaningful parts. The event segmentation theory proposed by \cite{Radvansky} is a new approach to how human cognition segments events. The event models presumed by the theory represent features of the current activity relevant to current goals and the models integrate information across the sensory modalities with information that may be more conceptual in nature.

The event segmentation theory entails that the representation of events involves biasing the pathway from sensory input to prediction. The theory says that the working models are disconnected from the sensory input and they store a static snapshot of the current event (preservation). This helps the event models overcome ambiguities and missing information. This process continues by comparing the predictions about the near future of an ongoing event with what actually happens, that is, monitoring the prediction error. If the prediction error suddenly increases, the event model will be updated by opening the inputs of the event models so that a new event is started. By opening the gates to a new operating model, perceptual information interacts with stored knowledge representations building a new event representation and when it is constructed, the prediction error is decreased and the gate closes.

\cite{zacks} presents three empirical experiments that have tested their event segmentation theory. The experiments are performed to investigate the ways in which the body movements of an actor predict when an observer will perceive event boundaries. In these experiments, participants segmented the movies of daily activities performed by a single actor using a set of objects on a tabletop. The results show that movement variables were significant predictors of the segmentation. The observers were more sensitive to the movements of the individual body parts and the distance between them than to the relative speed and acceleration of the body parts with respect to each other.

The psychological approach in event segmentation process introduces by \cite{zacks} is based on the possibility to predict the forthcoming movements. In other words, the system needs to predict the possible future movement of the actor based on what has been observed so far. As long as the prediction fits with the incoming stream of movements, it is maintained. Otherwise the system predicts that a new action has begun. As an example, take the action of scratching the head. The observer tracks the movements of the actor from when the arm is lifted. If the hand approaches the head and touches it then the observer categorizes it as head scratching and when the hand moves back and leaves the head it is considered the end of the action.

\subsection{Computational models of action recognition}
\label{sect11}
Human action recognition methods are largely dependent on the input modalities. There are three different types of input modalities that represent the actions performed; the RGB (color images), depth maps and skeleton information. The space-time volumes, spatio-temporal features and trajectories have been utilized for action recognition through the color images in the earlier methods proposed by \cite{Schuldt,Dollar,Sun}.

The color based methods are sensitive to color and illumination variations and thus they have limitations in recognition robustness. With advent of RGB-D sensors, the action recognition methods were developed based on depth maps, which are insensitive to illumination changes, color variations and provide us with rich 3D structural information of the scene. In the holistic approaches the global features such as silhouettes and space-time information are extracted from depth maps like the methods proposed in \cite{Oreifej,Li,Liu}. Other approaches extract the local features as a set of interest points from depth sequence (spatio-temporal features) and compute a feature descriptor for each interest point like the methods proposed in \cite{Laptev1,Wang1,Wang2}.

The cost-effective depth sensors are then coupled with the real-time 3D skeleton estimation algorithm introduced by \cite{Shotton}. By extraction of the spatiotemporal features from the 3D skeleton information such as the relative geometric velocity between body parts, relative joint positions and joint angles in \cite{Yao}, the position differences of the skeleton joints in \cite{Yang-Xiaodong1} or the pose information together with differential quantities (speed and acceleration) in \cite{Zanfir} the body skeleton information in space and time is first described. Then the descriptors are coupled with Principle Component Analysis (PCA) or another classifier to categorize the actions.

Such methods for action recognition utilize the pre-segmented and labeled datasets of actions, while online recognition of actions is crucial in real-time experiments with unsegmented sequences of actions. Next there is a description of other attempts in the literature to design computational models for online action recognition.

A main approach for online action recognition is based on the sliding window. \cite{Jalal} present a method, which segments human depth silhouettes using temporal human motion information and obtains skeleton joints through spatiotemporal human body information. Then it trains the hidden Markov model with the code vectors of the multi-fused features to recognize the segmented actions. \cite{Vieira} proposed a visual representation of 3D action recognition by space-time occupancy patterns. The method focuses on classifying the extracted feature vectors (interest points) from depth sensors by support vector machine. 

In \cite{Ellis} the skeleton data of specific events are converted to a feature vector of clustered pairwise joint distances between the current frame, 10 previous frames and 30 previous frames. The feature vectors are sent to classifiers that categorize actions based on canonical body poses. The method proposed in \cite{Lv} uses a dynamic programming algorithm to segment and recognize actions simultaneously. Their method decomposes the high-dimensional 3D joint representation into a set of feature spaces where each feature corresponds to the motion of a joint or related multiple joints. A weak classifier based on the hidden Markov model is formed for each feature and these classifiers are combined by the multi-class AdaBoost algorithm.

Among the neural network based methods for online action recognition there are convolutional neural networks based systems and recurrent neural network based systems for action recognition. A multi region two-stream R-CNN model for detecting actions in the videos is proposed by \cite{Peng}, by which the motion region network generates proposals complementary to those of an appearance region proposal network. They claim that stacking optical flow over several frames significantly improves frame-level action detection. A model of segment based 3D Convolutional Network is used for action localisation in long videos (see \cite{Shou}), which identifies candidate segments in a long video that may contain actions. A classification network learns action classification model to initialise the localisation model, which fine-tunes the learned classification network to localise an action instance. The UntrimmedNet model proposed by \cite{Wang-Limin} is composed of two main components implemented with feed-forward networks. The classification module and the selection module. They learn the action model from the video input and reason about the temporal duration of the action instances.

Among the recurrent neural network based system for action recognition is the method proposed by \cite{Singh}, a tracking algorithm is used to locate a bounding box around the performer in the video frames, which makes a frame of reference for appearance and motion and two additional streams are trained on motion and appearance. The pixel trajectories of a frame are utilised for the motion streams and a multi stream CNN is followed by a bi-directional Long Short-Term Memory (LSTM) layer to model long term temporal dynamics within and between the actions. The proposed model by \cite{Dave} proposes an action detection model for video processing, which utilises a series of recurrent neural networks that sequentially make top-down prediction of the future and later correct the predictions with bottom-up observations. The proposed approach by \cite{Ma} argues that when training the recurrent neural network and specifically a Long Short Term Memory (LSTM) model the detection score of the correct activity category or the detection score between the correct and incorrect categories should be monotonically non-decreasing as the model observers more of the activity. Therefore their model suggest the design of ranking losses to penalize the model on violation of such monotonicities, which are used together with classification loss in training of LSTM models. Finally the model of \cite{Li-Yanghao} proposes a joint classification regression recurrent neural network for online human action recognition from 3D skeleton data. The model applies the deep Long Short Term Memory (LSTM) subnetwork to capture the complex long range temporal dynamics and avoid the sliding window.

Among other neural networks for action recognition are ones proposed by \cite{Parisi1,Parisi2}. The method in \cite{Parisi1} proposes a neurobiologically motivated approach for noise-tolerant action recognition in real time. Their system first extracts pose and motion features of the action obtained from depth maps video sequences and later classifies the actions based on the pose motion trajectories. A two pathways hierarchy of growing when required (GWR) networks process pose motion features in parallel and integrate action cues to provide movement dynamics in the joint feature space. Then the GWR implementation is extended with two labelling functions to classify the action samples into the action categories. In another study  \cite{Parisi2}, Parisi et al., proposed deep neural network self-organisation for life-long action recognition. The system utilises a set of hierarchical recurrent networks for unsupervised learning of action representations with increasingly spatiotemporal receptive fields instead of handcrafted 3D features. The growth and the adaptation of the recurrent networks are driven by their ability to reconstruct temporally ordered input sequences and this makes the lifelong learning possible for the system. The visual representation obtained from unsupervised learning are associated with the action labels to satisfy the classification purposes.

This article presents instead a biologically inspired cognitive architecture that categorizes an ongoing action in an online mode. This means that the system receives information about an ongoing event such as body postures or object trajectories and continuously analyses the incoming data in order to categorize the action performed. To this end, the system needs to be capable of making an automatic segmentation together with categorization while different actions are sequentially performed. This is in contrast to the methods proposed by \cite{Wang-Pichao,Parisi1,Parisi2,Liu,Hou,Ijjina}, which rely on pre-segmented datasets of actions.

On the contrary to the methods presented by \cite{Ellis,Vieira,Lv}, the approach proposed in this article does not utilize a memory to preserve any previous frames since a trained SOM can connect consecutive features and as a result determines whether the coming frames belong to a particular action or not.

In contrast to the deep neural network based approaches for online action recognition \cite{Weinzaepfel,Peng,Shou,Singh,Ma,Dave}, which utilize 2D RGB images sensitive to illumination variations, color and texture changes, the method presented in this article uses skeleton data robust to scale and illumination changes and provides us with rich 3D structural information.

Next it comes with a description of how the biologically inspired cognitive architecture proposed in this article performs online recognition of actions. A more thorough comparison of the approach proposed in this article with other related methods in the literature is available in Section 4. 

\subsection{Proposed approach for online action recognition} 
One can view a particular event as being composed of a number of key components so that when the components are presented to the system in the right order, it can correctly categorize the event. As an example, consider the event of drinking a cup of coffee. In this case the key components could be ordered as follows: the hand reaches the cup, lifts it up, brings it to the mouth and then puts it down. Based on situational factors including where the cup is located (on the table, on the floor, etc), the properties of the cup (size, weight, shape, etc) and who the actor is (gender, age, physical condition, etc) the details of the ordered key components of the event will vary. The difference between the instances lies in how these components are combined to complete the event. The categorization of the action can be a function of kinematic factors such as position, speed, acceleration and the rate of performing a particular event.  

It seems that to solve the problem the system needs to learn the occurrence of forthcoming key components. For instance, take the earlier example of head scratching. The system tracks the movements of the actor that starts with lifting the arm. At this stage, more than one possible forthcoming action can be predicted e.g, head scratch, high arm wave, look at watch, forward punch, etc. Since there is more than one possible categorization, the system requires more information (key components) of the action performed to make a final decision. When the forthcoming movements fit more with the key components of initially possible actions,  for example, touching the head then the observer can more confidently categorize the action. 

Using the key frames  is also proposed for semantic segmentation in \cite{Li-Ruihao} in order to reduce the computational burden for video streams and improve the real-time performance. In their approach the convolutional neural network is utilized with spatial stream represented by images and temporal stream represented by image differences as their inputs.     

In this article, a cognitive architecture based on hierarchical self-organizing maps (SOM) consisting of two-layer SOMs together with a one-layer supervised neural network is used. The system contains a pre-processing unit consisting of ego-centered coordinate transformation, scaling and attention mechanism. The first-layer SOM is used to extract the features of each sequence of an action through observation of the pre-processed input data from a Kinect camera. The features are presented as the activation of neurons in the first-layer SOM. The key activations representing the actions are segmented by using a sliding window of fixed size and transferred to the second-layer SOM in order to cluster the second map into action categories. Finally, the third layer labels the categories that are formed in the second SOM and outputs the action names.    

Here the SOM is used for both feature extraction and pattern classification. Using the three layers of neural network in the hierarchical action recognition architecture introduces an online semi-supervised learning model \cite{Ding}, which resembles the human learning process in which the training samples are often obtained successively. In this way the observations arrive in sequence and the corresponding labels are presented very sporadically.

The main contributions of this article are listed as following: \\
(1) This article proposes a novel approach for online recognition of unsegmented action sequences inspired by humans's event perception and segmentation. \\
(2) The proposed approach is developed in a hierarchical cognitive architecture for action recognition. Different layers of the architecture are inspired by the biological organisms such as the pre-processing layer and the SOM layers.\\
(3) Although the performance of the system drops slightly in online experiments, the system remains highly accurate in performing the task online, which is more plausible and practical in real case scenarios.

\begin{figure*}
	\includegraphics[width=1.00\textwidth]{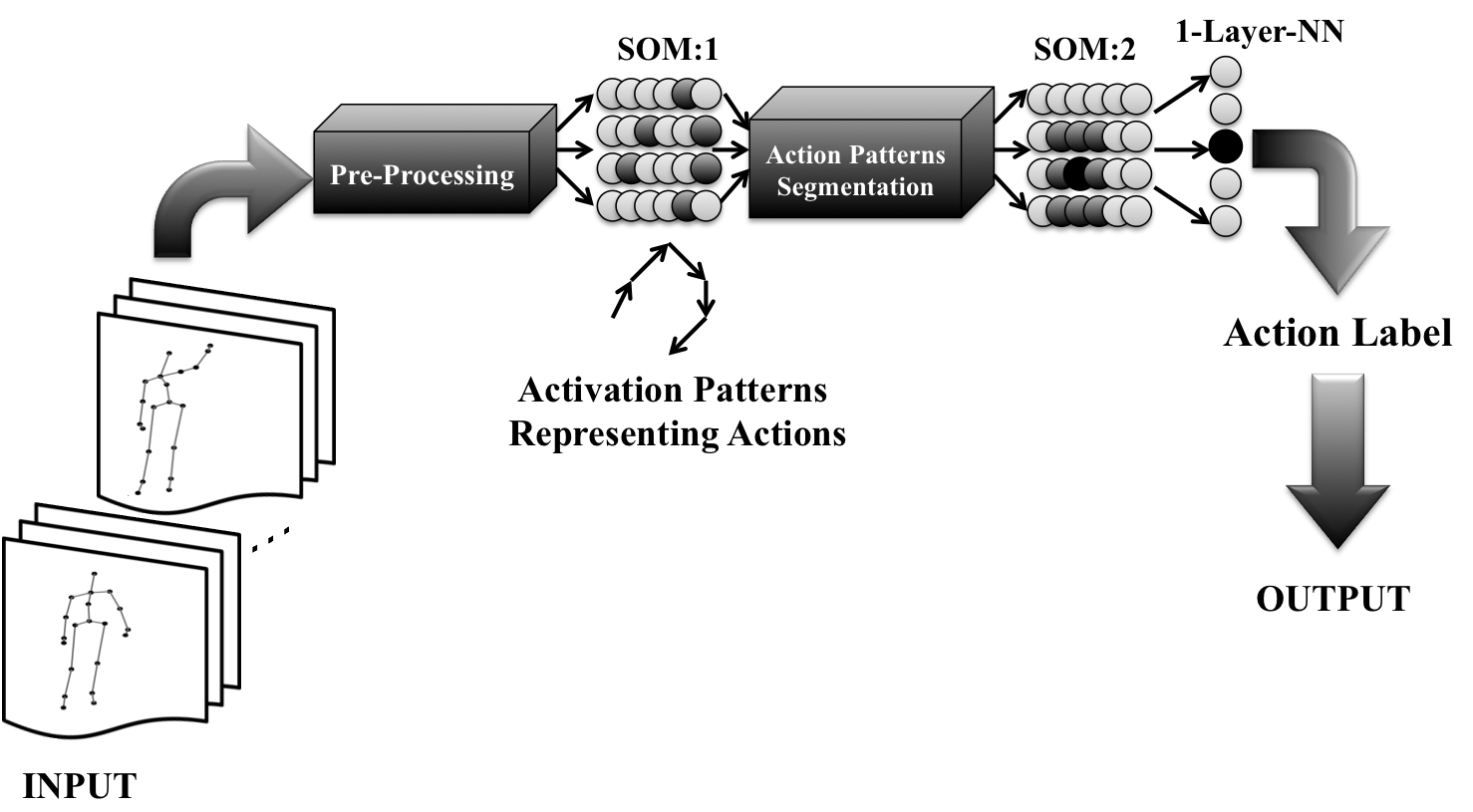}
	\caption{The three-layer action recognition architecture. The first and second layers consist of a SOM and the third layer is one-layer supervised neural network. The darker activation in the first SOM represents the activity trace during an action, which is also shown as a patterns made of arrows. The darker activation in the second SOM shows an example of a clustered region belonging to an action with stronger activation effect, and in the center of the region is the most activated neuron of the whole map, shown in black. The third layer (1-Layer-NN) is composed of the same number of neurons as the number of actions and the darker activation shows the action that the system recognized.}
	\label{fig:SOM_SOM_OutputLayer}      
\end{figure*}

\section{Methods}
\label{sec:3}
The multi-layer architecture shown in Fig.~\ref{fig:SOM_SOM_OutputLayer} is composed of several processing layers. The following section describes the implementations of the main layers such as preprocessing layer, SOM layers and one layer supervised neural network. More explanations regarding each layer are available in the earlier works \cite{Gharaee4,Gharaee5}.  

\subsection{Basic hierarchical SOM architecture}
\paragraph{Preprocessing:}The input data of an action performer is transformed into an ego-centered coordinate system located in the joint stomach. The 3D information of the joints right hip, left hip and stomach are used to build the ego centered coordinate system and all skeleton joints 3D information are transformed into this new coordinate system in order to compensate for having different viewing angle in relation to the Kinect camera. A more detailed description of the ego centered coordinate transformation is presented in \cite{Gharaee4}.  

To compensate for the different distances to the Kinect camera a scaling function is also applied to the input data. By transforming the skeleton postures into a standard size, the representations of the actions performed by the actor remain invariant of its different distances to the Kinect camera and as a result are set to a standard size.

Finally, an attention mechanism is applied to the input data in order to extract the parts of the body that are most active. The attention mechanism used in this architecture is inspired by human behavior, paying attention to the most salient parts of a scene, which in this case is the part of the body that moves the most during a particular action. To extract the active joints while performing an action the joints velocity is utilized and the attention mechanism selects the four most moving joints. As a result, by reducing the dimensionality of the input data in this way,  processing power and time is saved. 

\paragraph{SOM-Layers:}A SOM consists of an $I\times J$ grid of neurons with a fixed number of neurons and a fixed topology. Each neuron $n_{ij}$ is associated with a weight vector $w_{ij}\in{R}^n$ having the same dimension $K$ as the input vector $x(t)$. Each element of the weight vector is represented by three dimentions, $i$, $j$ and $k$. Where $0 \leq {i} < I$, $0 \leq {j} < J$, ${i}, {j}\in{N}$ represent the corresponding row and column of a neuron in the grid and $0 \leq {k} < K$ is equal to the input dimention. For  a squared SOM with equal number of rows and columns the total number of neurons is $N\times N$. All elements of the weight vectors are initialized by real numbers randomly selected from a uniform distribution between 0 and 1.

At time $t$, each neuron $n_{ij}$ receives the input vector $x(t)\in{R}^n$.
The net input $s_{ij}(t)$ at time $t$ is calculated using the Euclidean metric:

\begin{equation}
s_{ij}(t)=||x(t) - w_{ij}(t)||
\end{equation}

The activity $y_{ij}(t)$ at time $t$ is calculated by using the exponential function for each neuron of the grid:

\begin{equation}
y_{ij}(t)=e^{\frac{-s_{ij}(t)}{\sigma}}
\end{equation}

The parameter $\sigma$ is the exponential factor set to ${10^6}$. The role of the exponential function is to normalize and increase the contrast between highly activated and less activated areas.

The neuron $c$ with the strongest activation or the winner is selected because it represents the most similarity to the input vector. The weight vectors of all neurons $w_{ij}$ are adapted by using a Gaussian function centered at a winner neuron, $c$:

\begin{equation}
c=\mathrm {arg} \mathrm{ max}_{ij}y_{ij}(t).
\end{equation}

\begin{equation}
w_{ij}(t+1)=w_{ij}(t)+\alpha(t)G_{ijc}(t)[x(t)-w_{ij}(t)]. 
\end{equation}

The term $0 \leq \alpha(t) \leq 1$ shows the adaptation strength in which $\alpha(t) \rightarrow 0$ when $t \rightarrow \infty$. The neighborhood function $G_{ijc}(t)=e^{-\frac{||r_c - r_{ij}||}{2\sigma^2(t)}}$ is a Gaussian function decreasing with time, and $r_c \in R^2$ and $r_{ij} \in R^2$ are location vectors of neurons $c$ and $n_{ij}$ respectively. All weight vectors $w_{ij}(t)$ are normalized after each adaptation. Thus the winner neuron receives the strongest adaptation and the adaptation stength decreases by increasing distance from the winner. As a result the further the neurons are from the winner the more weakly their weights are updated. 

\paragraph{Output-Layer:}The output layer of the architecture is one-layer supervised neural network, which receives the activity traces of the second-layer SOM as the input vector with length $L$. The length $L$ is determined by the total number of neurons of the second-layer SOM. The output layer consists of a vector of $N$ number of neurons and a fixed topology. The number $N$ is determined by the number of action categories. As an example in the first experiment of this article the number of neurons of the output layer is set to 10, which is the number of actions categories.

Each neuron $n_{i}$ is associated with a weight vector  $w_{i}\in{R}^n$ and each element of the weight vector is represented by two dimensions $0 \leq {i} < N$, $0 \leq {l} < L$. Where All the elements of the weight vector are initialized by real numbers randomly selected from a uniform distribution between 0 and 1, after which the weight vector is normalized, i.e. turned into unit vectors.

At time \textit{t} each neuron $n_{i}$ receives an input vector $a(t)\in{R}^n$. The activity $y_{i}$ in the neuron $n_{i}$ is calculated using the standard cosine metric:

\begin{equation}\label{eq:7}
y_{i}=\frac{a(t)\cdot w_{i}(t)}{||a(t)||||w_{i}||}
\end{equation}

During the learning phase the weights $w_{i}$ are adapted by

\begin{equation}
w_{i}(t+1)=w_{i}(t)+\beta a(t)[y_{i} - d_{i}]
\end{equation}

The parameter  $\beta$ is the adaptation strength and $d_{i}$ is the desired activity for the neuron $n_{i}$.

%

\begin{figure*}
	\includegraphics[width=1.00\textwidth]{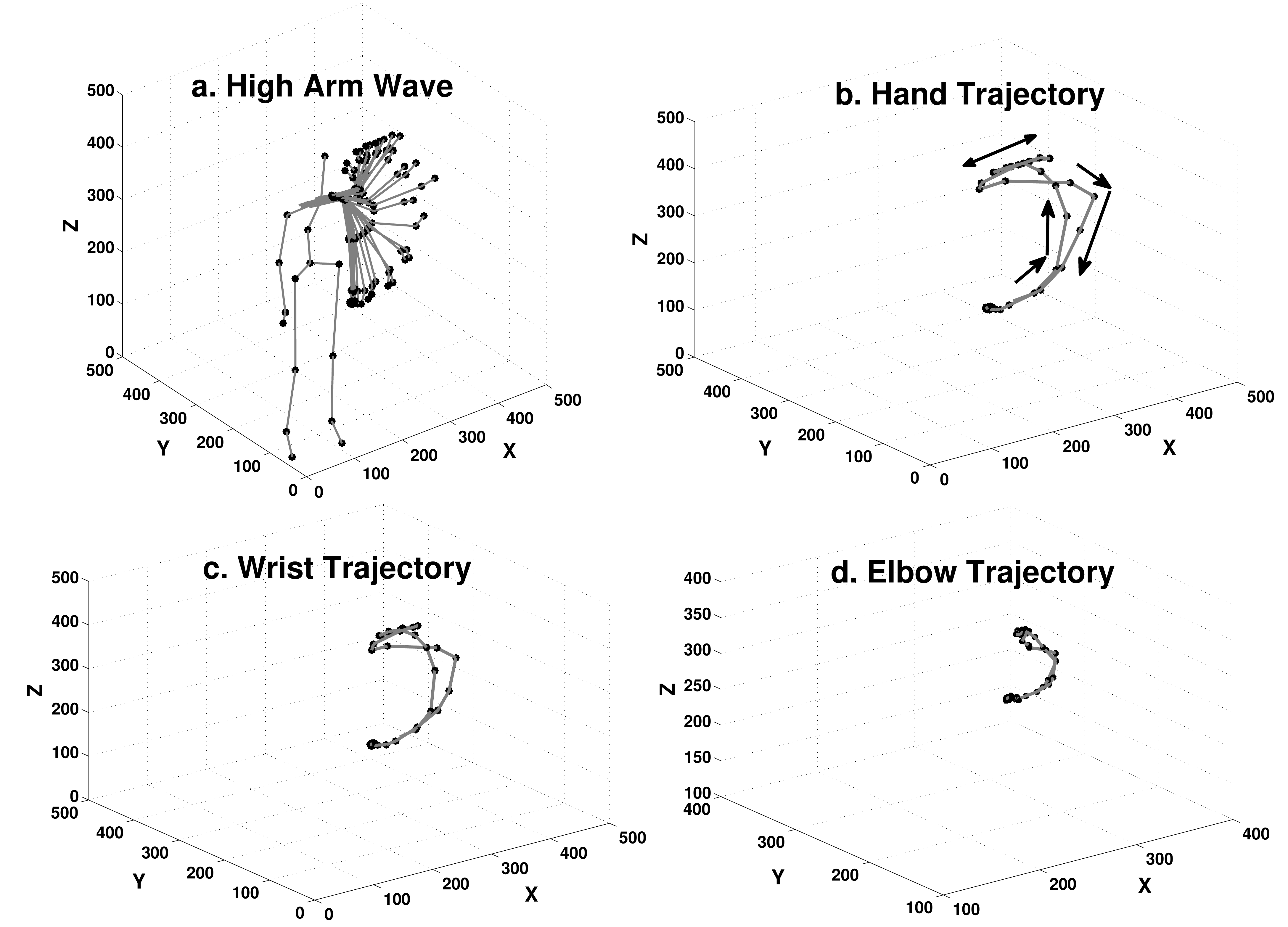}
	\caption{The 3D spatial trajectories of the body parts involved the most in performing the action \textit{high arm wave}. The performed action with the left arm is shown in part a as consecutive body postures. The spatial trajectory of the hand joint with directions of the movements is shown in part b. The spatial trajectories of the wrist and elbow joints are shown in parts c and d respectively.}
	\label{fig:trajectory}       
\end{figure*}

\subsection{Action pattern segmentation}
\label{sec:seg}
This section describes the technique utilized in hierarchical SOM architecture to implement the system for online real-time experiments with unsegmented input data of action samples. The module implemented to apply this technique receives the output vector patterns of the first-layer SOM and creates the input signal to the second-layer SOM. To this end it extracts the activations of the first SOM, which are elicited as a result of receiving the key postures of an ongoing action sample as the input of the system. In this way the output pattern vectors of the first-layer SOM corresponding to the input action samples are segmented. Thus the segmentation occurs automatically one step further into the system where the action feature vectors are created and segmented instead of input posture frames.

Each action sequence is represented by the the consecutive posture frames while each posture frame is composed of the 3D skeleton joints positions. The consecutive posture frames are applied to the system as the input.

The kinematics of actions are determined by the spatial trajectory of human skeleton components (like the joints) during the time interval the action performs. Temporal features are specified by the length and the order. The length is represented by the time interval during which the action performance is completed and the temporal order is represented by the sequential orders of the movements. As an example the action \textit{high arm wave} is performed by the left arm of the actor as represented in Fig.~\ref{fig:trajectory} part a. The left arm is composed of the joints left shoulder, left elbow, left wrist and left hand. The 3D temporal characteristics of these joints are presented in Fig.~\ref{fig:trajectory} parts b, c and d. As shown there, the hand and the wrist have almost similar spatial trajectories throughout time but on different scales (the smaller one is for the wrist and the larger one is for the hand). The elbow has a much smaller movement during acting compared to the hand and the wrist. The shoulder movement is even more limited and is thus not presented in the Fig.~\ref{fig:trajectory}.

Both the kinematic and dynamic characteristics of the action are crucial for perceiving it and they introduce the spatiotemporal features of the action. There are actions distinguished from one another only by one of these characteristics. For example the actions \textit{lift up} and \textit{put down} have similar posture frames of the arm movements representing their kinematics but with completely reversed temporal order. Therefore, the temporal order of the posture frames are the discriminating factor for these two actions.

On the other hand if an action is seen as a number of key components for example in the action \textit{horizontal arm wave} these key components can be lift the arm up, move the arm to the left/right direction, move the arm back to the reverse direction (right/left) and put the arm down. Based on the speed of performing the action each component can contain a number of posture frames that are similar. 

The spatiotemporal trajectory of an action extracted from consecutive 3D body postures of that action are received by the first-layer SOM and they activate specific areas of the map representing the input space. Pattern vectors are formed by connecting these ordered activations of the performing action. Let’s assume that there is a distinct elicited activation for each posture of an action sequence. Then as a result there will be a series of elicited activations for the whole action sequence. So the key components of the action can relate to the key postures as well as the key elicited activations in the SOM. 

\begin{figure*}
	\includegraphics[width=0.90\textwidth]{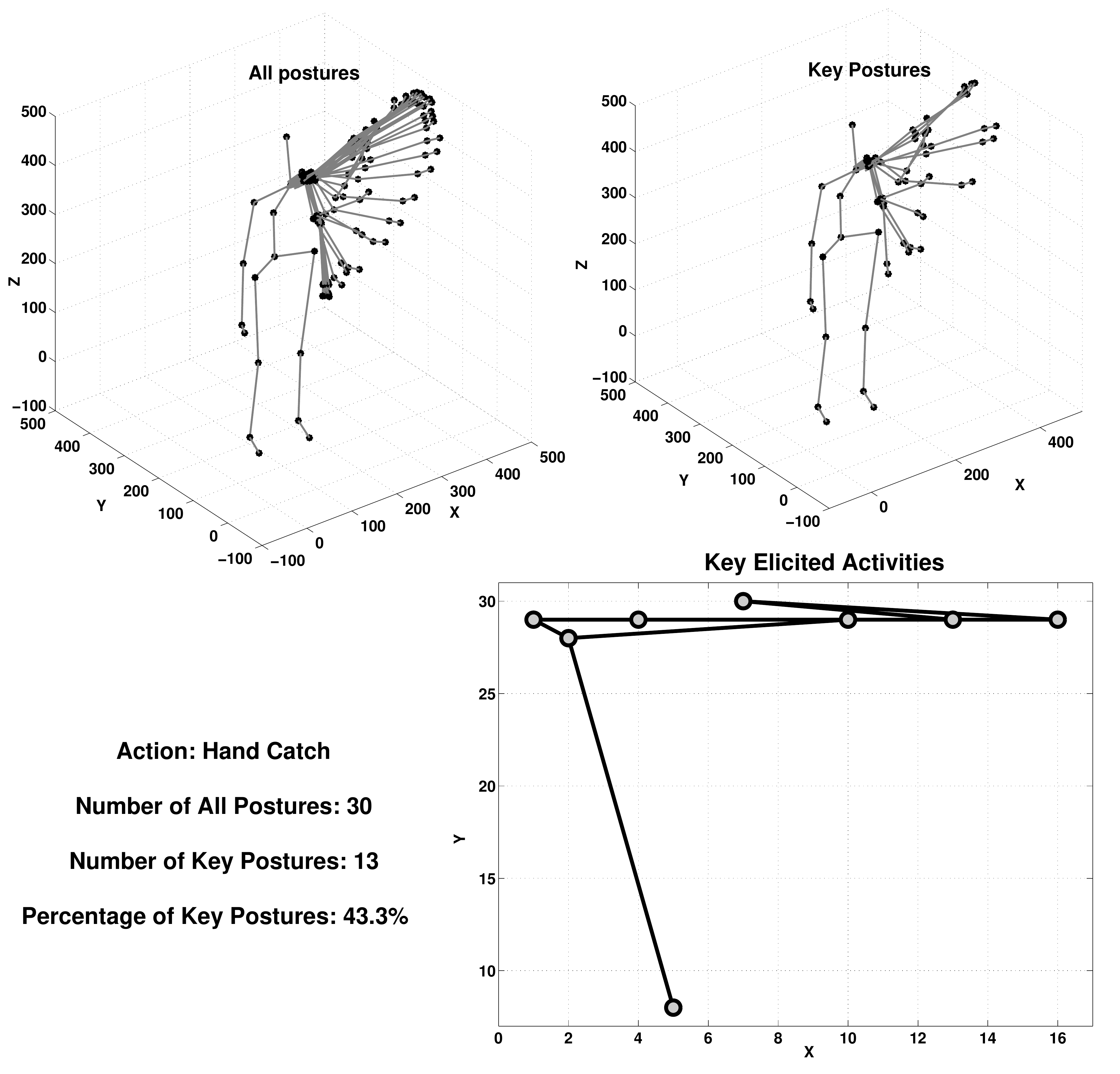}
	\caption{The top left of the figure shows all consecutive posture frames while performing the action \textit{hand catch}. The top right part of the figure shows only those postures with unique activation in the first-layer SOM extracted by the action patterns segmentation unit and the bottom right part of the figure shows the action pattern as a result of key activations elicited in the first-layer SOM. In the bottom right part of the figure the activated neurons are represented by circles and some of them are activated more than once in non-consecutive iterations.}
	\label{fig:allVskeyPostures}       
\end{figure*}

Thus action sequences are segmented by extracting and segmenting the key activations in the first-layer SOM. In the left top part of Fig.~\ref{fig:allVskeyPostures}, all consecutive postures of the action \textit{hand catch} are shown and in the right top part those postures with key activations in the first-layer SOM (key postures) are depicted. The right bottom part of Fig.~\ref{fig:allVskeyPostures} shows the key activations of the SOM (action pattern) corresponding to the same action sample. 

The same action can have a completely different visual appearance. Variations can occur because of performing speed, clothing etc. Based on how an action is performed and the nature of the actions, the length of elicited key activations in the first-layer SOM will be different. If an action is performed too slowly the number of similar consecutive postures will increase, but each posture of an action sample is considered as a key posture if and only if it elicits a unique activity in the SOM.

Next there is a detailed description of how the proposed approach to recognizing unsegmented action sequences is developed in hierarchical SOM architecture. Let's start with input space of the actions, which is composed of action sequences:

\begin{equation} \label{eq:1}
input =\left \{s_{1}, s_{2}, s_{3}, ...,s_i, ..., s_{N_{s}} \right \}, 
\end{equation}

where $0< i <N_{s}$ and $N_{s}$ is the total number of action sequences of the dataset, which is 276 for the first experiment. Each action sequence $s_{i}$ is composed of the consecutive posture frames:

\begin{equation} \label{eq:2}
s_{i} =\left \{p_{1}, p_{2}, p_{3}, ...,p_j, ... , p_{N_{p}} \right \}, 
\end{equation}

where $0< j <N_{p}$ and $N_{p}$ is the total number of posture frames representing an action sequence and varies for different action sequences. A posture frame $p_{j}$ contains spatiotemporal features represented by 3D information of the skeleton joints:

\begin{equation} \label{eq:3}
p_{j} =\left \{d_{1}, d_{2}, d_{3}, ...,d_k, ..., d_{N_{d}} \right \}, 
\end{equation}

where $0< k <N_{d}$ and $N_{d}$ is the full dimension of spatiotemporal features. For 3D features of the skeleton joints positions, $N_{d}$ can represent the total number of identified joints in 3D. As an example if 20 joints are extracted from each posture frame then $N_{d}=20\times3=60$. 

After some preprocessing the consecutive spatiotemporal feature vectors represented by posture frames $p_{i}$ are received by the first-layer SOM. The activity traces of the first-layer SOM are extracted as 2D positions of the activated neurons. The consecutive elicited activities of each action sequence is: 

\begin{equation} \label{eq:4}
a_{i} =\left \{[x_{1},y_{1}]_{i}, [x_{2}, y_{2}]_{i}, ...,, [x_q, y_q]_i, ... [x_{L_{i}}, y_{L_{i}}]_i\right \}, 
\end{equation}

where $0< i <N_{s}$ and $[x_q , y_q]$ shows the location of a neuron on the 2D neural map. The term $0< q < L_{i}$ and $L_{i}$ shows the full length of a pattern vector varying for different action sequences. The consecutive activity pattern vector of all action sequences is:

\begin{equation} \label{eq:5}
\begin{aligned}
A =\left \{a_1,\quad a_2, \quad a_3, \quad ...,\quad a_i, ...,  \quad a_{N_{s}}\right \} .
\end{aligned}
\end{equation}

At time $t$, 2D locations of the elicited activation of first-layer SOM, $a_{in}(t)=[x_{t}, y_{t}]$, is received as the input of pattern vector segmentation layer. Based on how the actions are performed there are similar consecutive elicited activation representing similar posture frames of the action sequence. Such similar consecutive activity traces are first mapped into a unique activation. Then a constant length of segmentation $T$ is applied to segment the action sequences, which determines when the elicited activations representing the action starts/ends. The segmented vector, which is the result of receiving real-time action sequences is: 

\begin{equation} \label{eq:7}
\begin{aligned}
a_{out}= \{[x_{0}, y_{0}], [x_{1}, y_{1}],... [x_{T}, y_{T}], \\ [x_{T+1}, y_{T+1}], [x_{T+2}, y_{T+2}],...,[x_{2T}, y_{2T}], \\ [x_{2T+1}, y_{2T+1}], [x_{2T+2}, y_{2T+2}],...,[x_{3T}, y_{3T}] ,..., [x_{N_tT}, y_{N_tT}]\},
\end{aligned}
\end{equation}

where $T$ is a constant value used for segmentation and $N_t$ is the total number of segmented vectors. The segmentation size is calculated from the average length of the key activity traces for all action sequences of the training data set. The same size is applied to the activity traces of all action sequences in both training and test data set. The action pattern vectors should be segmented in a way to encompass key activations of all action sequences, so ideally it shouldn’t be too large to not contain the key activations of more than one action sequence and at the same time it shouldn’t be too small so that it does not ignore the key activations of a single action sequence.    

The size of segmentation is usually determined empirically in the experiments. By training the system for several trials the segmentation size might be updated. As an example in the first experiment of this article the longest and the shortest activity traces for a sub set of dataset contain 59 and 12 key activations respectively, and after some tuning the segmentation size $T$ is set to 30.

\section{Results}
\label{sec:exp}
The performance of the architecture shown in Fig.~\ref{fig:SOM_SOM_OutputLayer} is evaluated in two experiments. In these experiments two kinds of input data are used. First the architecture is tested on a publicly available dataset called MSR-Action3D dataset (\cite{MSR}). The MSR-Action3D is the first public benchmark RGB-D set collected by a Kinect sensor and it provides us with the skeleton data of the actions performed. For the second experiment a new dataset is collected, which is composed of 3D postures of a human actor performing actions using a Kinect sensor to validate the system in online experiments. 

The neural modeling framework Ikaros \cite{balkenius} has been used to implement the architecture and also to perform the experiments. The results were filmed and demo movies were created for both experiments. The movies are accessible on the web page \cite{Johnsson3}.

\subsection{Experiment 1}
In the first experiment the ability of the hierarchical SOM architecture to categorize actions is tested in an online mode by using a data set of actions composed of sequences of 3D joints postures. This dataset contains 276 samples with 10 different actions performed by 10 different subjects in 2 to 3 different events. Each action sample is composed of a sequence of frames where each frame contains 20 joint positions expressed in 3D Cartesian coordinates. The actions of the first experiment are: \textit{high arm wave}, \textit{horizontal arm wave}, \textit{using hammer}, \textit{hand catch}, \textit{forward punch}, \textit{high throw}, \textit{draw x}, \textit{draw tick}, \textit{draw circle}, \textit{tennis swing}. 

All action samples of the MSR-Action3D dataset are segmented and labeled with the names of the corresponding action, subject and event. Thus to run the first experiment in online real-time mode the system is provided with random selection of unsegmented data of consecutive actions as the input and is using the labeling information to validate the system performance by comparing whether the recognized action by the architecture correctly matches the real action performed.

\begin{figure*}
	\includegraphics[width=1.00\textwidth]{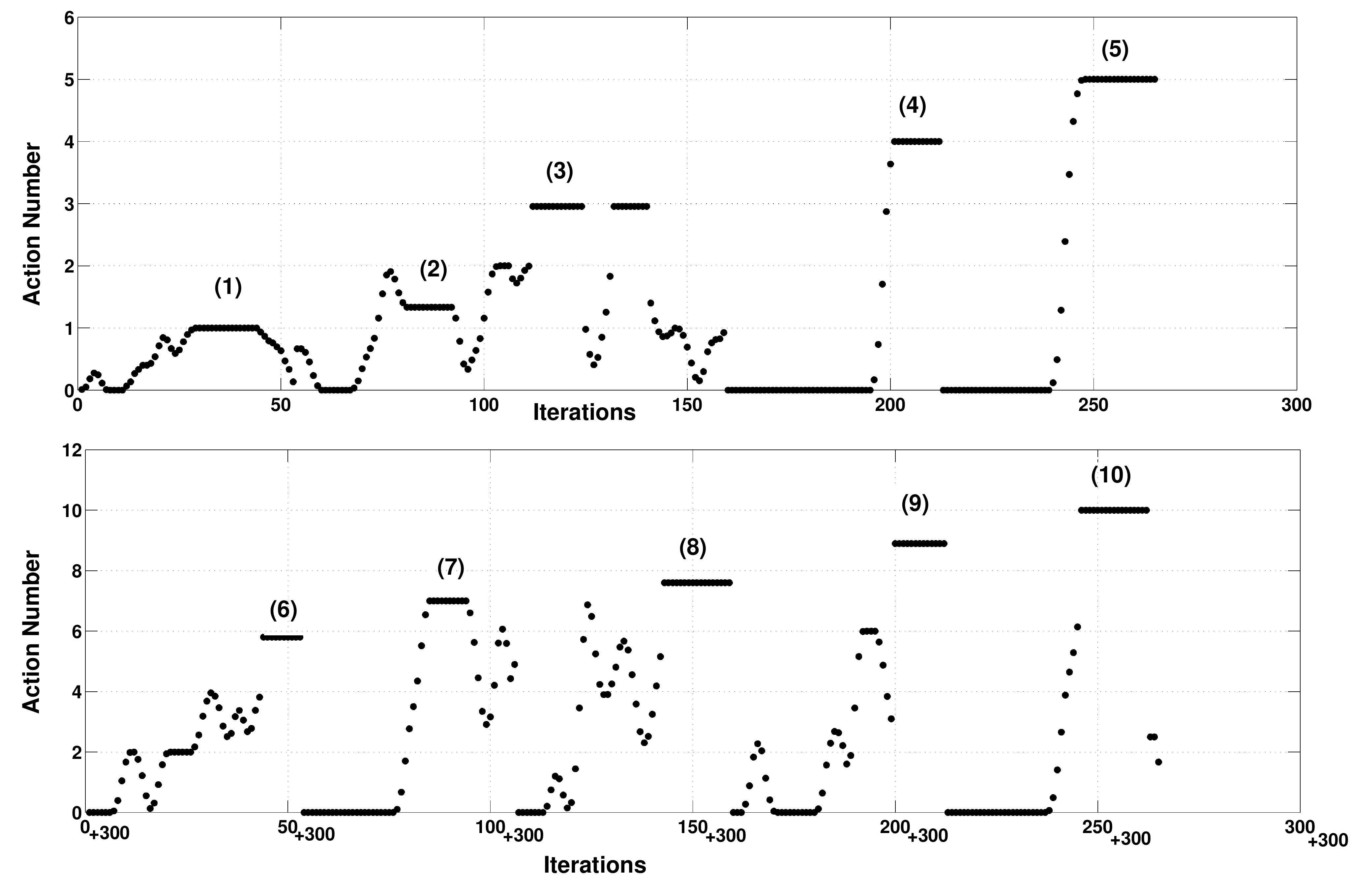}
	\caption{Online action recognition experimental results obtained through application of segmentation on MSRAction3D dataset. The upper row shows the recognition results of the actions: \textit{high arm wave (1)}, \textit{horizontal arm wave (2)}, \textit{using hammer (3)}, \textit{hand catch (4)}, \textit{forward punch (5)}, while the lower row depicts the recognition results of the actions: \textit{high throw (6)}, \textit{draw x (7)}, \textit{draw tick (8)}, \textit{draw circle (9)}, \textit{tennis swing (10)}. One iteration counts when a unique posture frame is received by the first-layer SOM. The average recognition accuracy corresponding to the action performed is calculated and multiplied by the action number and plotted.}
	\label{fig:SegResult}     
\end{figure*}

For this experiment, the dataset was split into a training set containing 80\% of the action instances randomly selected from the original dataset and a generalization test set containing the remaining 20\% of the instances. Then the neural network system was trained with randomly selected instances from the training set in two phases, the first to train the first-layer $30\times30$ neurons SOM and the second to train the second-layer $35\times35$ neurons SOM and the output-layer containing $10$ neurons.

In order to make the result invariant of the order of action sequences different random selections of test samples are applied to the trained system and the average categorization results of all test samples of each action was shown in Fig.~\ref{fig:SegResult}. This process repeated for different random selections of training and test samples from all action sequences.

As shown in Fig.~\ref{fig:SegResult}, the actions are correctly categorized already after a few iterations from when their input patterns are applied to the second SOM. One iteration counts when a unique posture frame is received by the first-layer SOM. Naturally, it takes some iterations for the input pattern to cover all the corresponding key activations as a result of first-layer SOM receiving key postures of the corresponding action. The correct categorization continues for several iterations and then it shifts to zero because of the updating process,which occurs in the activation of the neuron representing the correct action performs. During updating process, system starts building an input sequence pattern for a new action. In this experiment the average performance of 75\% correct categorization is obtained for the generalization test data when the segmentation technique is used.

\begin{figure*}
	\includegraphics[width=1.00\textwidth]{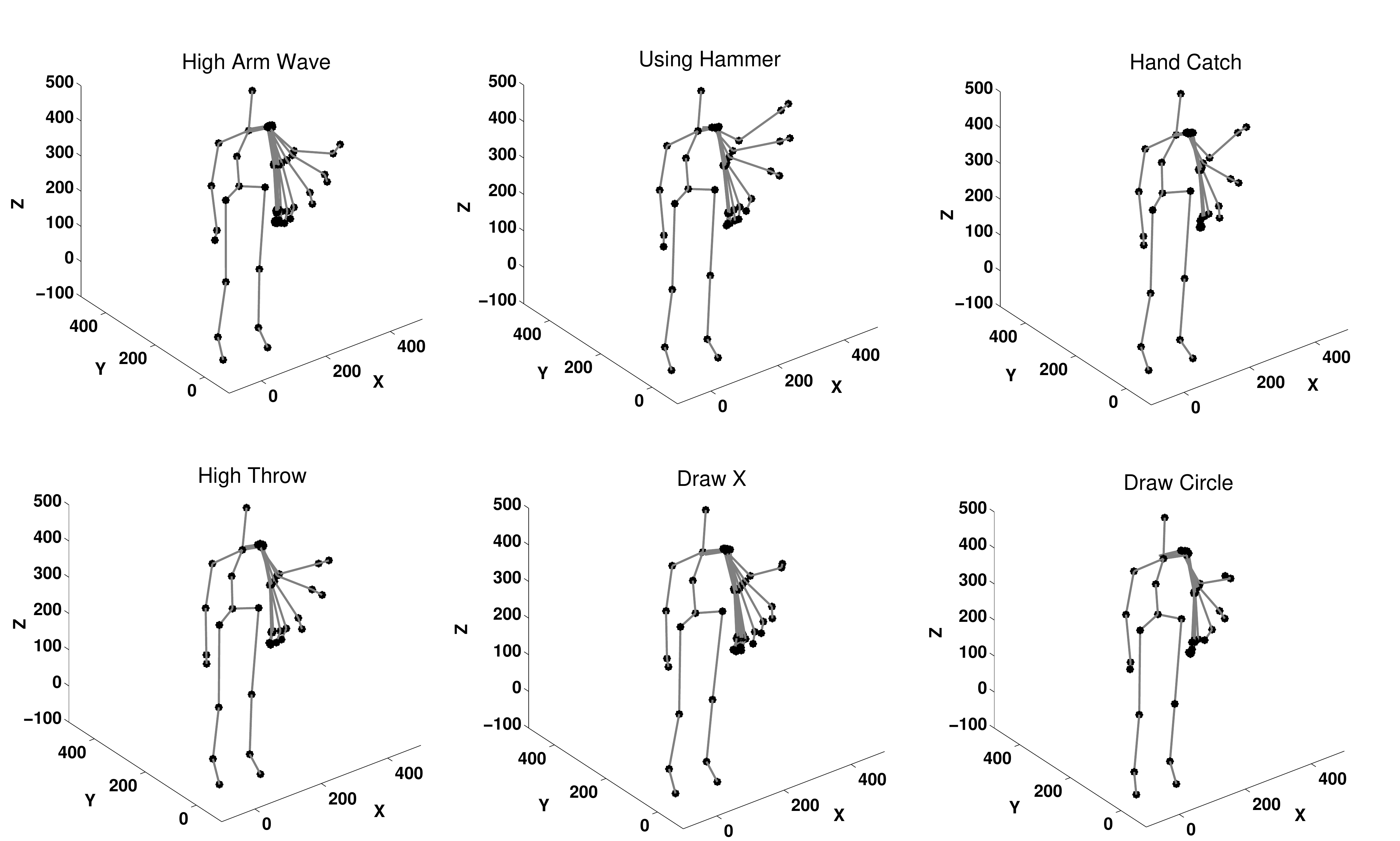}
	\caption{This figure shows the first postures of the six actions performed in order to represent the similarities of these postures that belong to distinct actions.}
	\label{fig:6act}       
\end{figure*}

A certainty measure is also used in which the online categorization of the action performed is given as output if it continues for a number of consecutive iterations, which is between 5 and 7 iterations (having around 80\% of the action input). This is done to achieve a more stable and robust recognition result. For this it is necessary that the sliding window contains the key activations of an action sequence continuously for a number of consecutive iterations, which is a restriction for the categorization task. The 75\% correct categorizations should be compared to 83\% correct that was obtained \cite{Gharaee3} when the data set was segmented in advance in an offline experiment. The results show that the performance drops when the segmented action pattern vector of fixed length is used but at the same time the system is capable of running online experiments.

What makes the categorization task difficult is when the actions have similar components. As an example take the actions \textit{high arm wave}, \textit{using hammer}, \textit{hand catch}, \textit{high throw}, \textit{draw X} and \textit{draw circle}. In all of these actions the first postures begin with lifting the arm up. It can be seen as the same movement although the actions are different. Fig.~\ref{fig:6act} shows a number of beginning postures of these actions and the plotted postures are similar even though they belong to distinct actions. Therefore the system should categorize them as different categories to distinguish them. The similarities also increase the delay of the system before it makes the right guess as to the action performed.

\subsection{Experiment 2}
In the second experiment, the online experiment presented in \cite{Gharaee2} is developed by implementing the segmentation technique proposed in this study. Then a data set of actions composed of 3D joints postures is collected by using a Kinect sensor. The actions of the second experiment are: \textit{high arm wave}, \textit{forward punch}, \textit{draw x}, \textit{draw circle} and \textit{tennis swing}.

In this experiment, the same architecture is used with similar preprocessing layer to the input data. The system is trained on a dataset containing 60 sequences of 5 different actions in which there are 12 different sequences of each action. The actions are performed by a single actor in 2 to 3 different events. 

After the system has been trained on the dataset with the result of 100\% correct categorization accuracy, an online real-time generalization test experiment is performed on the trained weights of the system in which the actor performed similar actions in front of a Kinect camera. The actions are selected randomly and performed in several trials. As a result of this experiment, an average accuracy of 88.74\% correct categorization for the generalization test experiments is obtained. 

The proposed segmentation technique makes it possible to run online experiments. It means that the system is capable of categorizing actions continuously as the actions are performed. This is different and more difficult compared to the condition when the system is tested in offline mode on pre-recorded data sets. For human-robot interaction and many other applications of action recognition, it is mandatory to be able to perform the categorization in an online mode.

In this study the aim has been to show the capacity of the SOM architecture in online implementations. By running the experiments online, the system accuracy drops slightly, as shown in Table \ref{table}, compared to when pre-segmented data are given, but the practicality of the online mode outweighs the drop.

\begin{table}
	\centering
	\caption {The performance of online action categorization task in the real-time experiments by segmenting action patterns.} \label{tab:title}
	\begin{tabular}{|c c c|} 
		\hline
		& Offline test experiment   & Online test experiment \\ [0.5ex] 
		& with segmented actions  & with unsegmented actions  \\
		\hline
		Experiment.1 & 83\% & 75\% \\ 
		\hline
		Experiment.2 & 94.29\% & 88.74\%  \\
		\hline
	\end{tabular}
	\label{table}
\end{table}

\section{Discussion}
In this article an action recognition task is performed in online real-time experiments. Therefore, the segmentation problem in dealing with the datasets of unsegmented action sequences needs to be solved. The segmentation problem addressed in this article is related to detection of the start and/or end of the action performed in a time series of consecutive action sequences.

The simple way is to manually segment the untrimmed videos of action sequences, which is highly expensive. Most of the methods such as \cite{Wang-Pichao,Parisi1,Parisi2,Liu,Hou,Ijjina} rely on pre-segmented datasets of actions and thus they are evaluated with respect to the benchmarks containing labeled action sequences.

Among the methods for online action recognition tasks such as STOP feature vectors in\cite{Vieira}, canonical poses in \cite{Ellis} and feature spaces of a joint or related multiple joints in \cite{Lv} there are certain features extracted first. Then these features are applied to a particular classifier to be categorized. An important question concerns to what degree the extracted features represent the action sequences in other words, how much information is lost in the data compression of the features. 

The feature extraction in some of these methods is performed for fixed time intervals, for example every 5 frames in \cite{Vieira} or between 10 and 30 previous frames in \cite{Ellis}. A limitation of this approach is that it requires having access to a certain number of frames for each iteration, which necessitates a capacity to preserve previous information and may also result in a delay in achieving results. 

In contrast to the methods proposed in \cite{Ellis,Vieira}, by learning the sequential relation of the consecutive posture frames, the approach proposed here is independent of allocation of memory to preserve any previous frames since a trained SOM can connect consecutive features through connecting consecutive activated neurons of the lattice, and as a result determine whether coming frames belong to a particular action or not.

The system produces the action label when it perceives a number of consecutive key frames, which is less than $T$ for the majority of the action sequences. There is no pre-set delay considered in the structure of the system and the delay occurs mainly due to the fact that the certainty of the system increases when more key features are observed. Because of this, there is less delay in obtaining a categorization response from the system. This aspect accords with human action categorization. As an example, when a person lifts up his arm he might want to look at his watch, scratch the head or wave to greet a friend. Thus one cannot recognize what he is doing until more key components of the action are received.

In the approach proposed in this article instead a neutral pause between actions is not employed as done in \cite{Vieira}, so the whole stream of actions is applied as the input of the system. Moreover, the allocation of both the start and the end of each action sequence could be critical and not only the start of the action, as has been proposed in \cite{Ellis}. The hierarchical SOM system addresses the problem of allocating the beginning and ending of the actions by learning the sequential relations between the consecutive frames so that when it receives two consecutive frames it can detect whether they belong to the same action or not.

Although the system proposed in \cite{Lv} is claimed to be capable of automatic recognition and segmentation of 3D human actions, it is not clear how this system works in online experiments because it is tested on a collected MoCap dataset of actions. In contrast, this study presents an online experiment by using a Kinect sensor in real time and the ability of the system is tested on new data of online actions.

The neural network based approaches for online recognition of actions proposed in \cite{Weinzaepfel,Peng,Shou,Singh,Ma,Dave} use the RGB images as the input modality. Although the RGB images provide input data with rich characteristics of shape, color and texture, they are 2D images sensitive to illumination variations, color and texture changes so the performance of the task is largely dependent on the quality of the input images. In fact it is quite expensive to produce and use high quality images of actions. Since this requirs expensive cameras for data collection and the dataset produced by these cameras contains high resolution images of large dimensionality, which necessiates more time and processing power for data analysis. Another limitation of these approaches is that most of deep learning methods rely on large labeled training data, which adds even more cost in running the experiments.

On the other hand the skeleton data is robust to scale and illumination changes and provides us with rich 3D structural information of the scene by calculating the positions of the human joints in 3D space as a high-level feature representing the kinematics and dynamics of the actions. Furthermore, skeleton data can be invariant to human body rotation and speed of the motions. Similar to the method proposed by this article the skeleton information is utilized in an online action detection approach based on joint classification regression recurrent neural network (see \cite{Li-Yanghao}). This method utilises the 3D joints input similar to the method proposed by this article. The proposed approach of \cite{Li-Yanghao} is tested on an input set of actions containing 10 different actions and obtain average recognition accuracy of 65\%. Although different types of action input are used to test the model proposed by \cite{Li-Yanghao} and the model proposed in this article, the proposed model of this article obtains better results. As shown in the result section my architecture obtains overall recognition accuracy of 75\% and 88.74\% in online test experiments on two action sets containing 10 and 5 different actions.

The proposed architecture here is capable of recognizing actions in online experiments. The system extracts key features of each action sequence, represented as a pattern vector, and uses the learned vector as the representative of that action sequence. First-layer SOM of the architecture learns the consecutive postures of actions and extracts the action patterns while the second SOM classifies the segmented patterns into action categories. To increase the categorization certainty it is checked whether the system sends the same action as its output during a few iterations and, if so, that action is considered as the output of the system. By using this method the performance accuracy drops slightly compared to pre-segmented data but the certainty of the system is improved.

Using cognitive mechanisms such as attention makes the system more biologically plausible. Since the attention mechanism is inspired by human behavior, paying attention to the most salient parts of a scene, which in this case is the part of the body that moves the most during a particular action. As a result, by reducing the dimensionality of the input data in this way,  processing power and time is saved and at the same time the performance of the action recognition system significantly increased (see also\cite{Gharaee1}).  

It should be mentioned that in the approach presented in this article, the system is trained on limited dataset of labeled action sequences and it is able to generalize, i.e. the network can recognize or characterize input it has never encountered before.

The results of the experiments performed in this study show that the recognition of unsegmented actions in online test experiments is quite high. When the performance results of this paper is compared to our earlier empirical studies, whether they are online experiments \cite{Gharaee2,Gharaee5} or offline experiments \cite{Gharaee3,Gharaee4}, there is a decrease in the acquired recognition accuracy of the system. An explanation for this may be that different sequences of actions span different time intervals, while a constant cut-off length is allocated to all of them through the sliding window.

One limitation of using 3D skeleton data is with the reduction in accuracy due to the environmental noise and the transformation of different modalities. Another limitation with the method proposed in this article is with setting the size of segmented action pattern vectors. It requires computational effort to find out the best size for action pattern segmentation for the training data. On the other hand if the test actions performed too different from the training samples, the recognition accuracy might drops.

\section{Conclusion}
In summary this article proposes a new method for human action recognition and segmentation using a SOM-based system. The hierarchical architecture consists of two-layer SOMs together with one-layer supervised neural network. The system is validated on different experiments. First the system is tested on the MSR-Action3D dataset and then it is validated on a dataset collected by a Kinect sensor in online experiments.  

In order to improve action recognition and segmentation performance, it is planned plan to design and implement a sliding window of the pattern vectors with variable size that is adapted to the actual size of the key activations of the corresponding action sequence. Another plan would be to develop a method for solving the segmentation problem that calculates the prediction error between the actual forthcoming movement of the actor and the predicted one by the system while an action is performed and use this error value to determine when the action ends and a new action begins. To this end the associative SOM (see \cite{Hesslow} and \cite{johnsson1}) for action recognition and segmentation could be used.

\section*{Acknowledgment}
This work was partially supported by the Wallenberg AI, Autonomous Systems and Software Program (WASP) funded by the Knut and Alice Wallenberg Foundation.

\bibliography{References}

\newcommand{\noop}[1]{}
\begin{thebibliography}{10}
\expandafter\ifx\csname url\endcsname\relax
  \def\url#1{\texttt{#1}}\fi
\expandafter\ifx\csname urlprefix\endcsname\relax\def\urlprefix{URL }\fi
\expandafter\ifx\csname href\endcsname\relax
  \def\href#1#2{#2} \def\path#1{#1}\fi

\bibitem{Gharaee2}
Z.~Gharaee, P.~G\"ardenfors, M.~Johnsson, Action recognition online with
  hierarchical self-organizing maps, in: Proceedings of the International
  Conference on Signal Image Technology and Internet Based Systems(SITIS),
  2016, dOI: 10.1109/SITIS.2016.91.

\bibitem{Gharaee3}
Z.~Gharaee, P.~G\"ardenfors, M.~Johnsson, Hierarchical self-organizing maps
  system for action classification, in: Proceedings of the International
  Conference on Agents and Artificial Intelligence (ICAART), 2017, dOI:
  10.5220/0006199305830590.

\bibitem{Gharaee4}
Z.~Gharaee, P.~G\"ardenfors, M.~Johnsson, First and second order dynamics in a
  hierarchical som system for action recognition, Applied Soft Computing 59
  (2017) 574--585, dOI: https://doi.org/10.1016/j.asoc.2017.06.007.

\bibitem{Gharaee5}
Z.~Gharaee, P.~G\"ardenfors, M.~Johnsson, Online recognition of actions
  involving objects, Biologically Inspired Cognitive Architectures (BICA) 22
  (2017) 10--19, dOI: https://doi.org/10.1016/j.bica.2017.09.007.

\bibitem{Gharaee6}
Z.~Gharaee, Action in Mind: A Neural Network Approach to Action Recognition and
  Segmentation, Lund University: Cognitive Science, 2018.

\bibitem{Michotte}
A.~Michotte, The Perception of Causality, Basic Books, 1963.

\bibitem{Gibson1}
J.~J. Gibson, The Senses Considered as Perceptual Systems, Oxford, England:
  Houghton Mifflin, 1966.

\bibitem{Gibson2}
J.~J. Gibson, The Ecological Approach to Visual Perception, Hillsdale, NJ:
  Lawrence Erlbaum, 1979.

\bibitem{johansson}
G.~Johansson, Visual perception of biological motion and a model for its
  analysis, Perception \& Psychophysics 14~(2) (1973) 201--211.

\bibitem{Gardenfors1}
P.~G\"ardenfors, Representing actions and functional properties in conceptual
  spaces, in: Body, Language and Mind, Vol.~1, Mouton de Gruyter, Berlin, 2007,
  pp. 167--195.

\bibitem{Gardenfors2}
P.~G\"ardenfors, M.~Warglien, Using conceptual spaces to model actions and
  events, Journal of Semantics 29 (2012) 487--519.

\bibitem{Radvansky}
G.~A. Radvansky, J.~M. Zacks, Event Cognition, Oxford University Press, 2014.

\bibitem{zacks}
J.~M. Zacks, S.~Kumar, R.~A. Abrams, R.~Mehta, Using movement and intentions to
  understand human activity, Cognition 112 (2009) 201-- 216,
  dOI:10.1016/j.cognition.2009.03.007.

\bibitem{Schuldt}
C.~Schuldt, I.~Laptev, B.~Caputo, Recognition human actions: a local svm
  approach, in: Proceedings of IEEE International Conference on Pattern
  Recognition, Vol.~3, 2004, pp. 32--36, dOI: 10.1109/ICPR.2004.1334462.

\bibitem{Dollar}
P.~Dollar, V.~Rabaud, G.~Cottrell, S.~Belongie, Behavior recognition via sparse
  spatio-temporal features, in: IEEE International Workshop on Visual
  Surveillance and Performance Evaluation of Tracking and Surveillance, 2005,
  pp. 65--72, dOI: 10.1109/VSPETS.2005.1570899.

\bibitem{Sun}
J.~Sun, X.~Wu, S.~Yan, L.-F. Cheong, T.-S. Chua, J.~Li, Hierarchical
  spatio-temporal context modeling for action recognition, in: IEEE Conference
  on Computer Vision and Pattern Recognition (CVPR), 2009, pp. 2004--2011,
  dOI:10.1109/CVPR.2009.5206721.

\bibitem{Oreifej}
O.~Oreifej, Z.~Liu, Hon4d: Histogram of oriented 4d normals for activity
  recognition from depth sequences, Proceedings of the IEEE Conference on
  Computer Vision and Pattern Recognition (CVPR)DOI: 10.1109/CVPR.2013.98
  (2013).

\bibitem{Li}
W.~Li, Z.~Zhang, Z.~Liu, Action recognition based on a bag of 3d points, in:
  IEEE Computer Society Conference on Computer Vision and Pattern Recognition
  Workshops (CVPRW), 2010, pp. 9--14, dOI:10.1109/CVPRW.2010.5543273.

\bibitem{Liu}
M.~Liu, H.~Liu, C.~Chen, Enhanced skeleton visualization for view invariant
  human action recognition, Pattern Recognition 68 (2017) 346--362,
  dOI:https://doi.org/10.1016/j.patcog.2017.02.030.

\bibitem{Laptev1}
I.~Laptev, On space-time interest points, International Journal of Computer
  Vision 64 (2005) 107--123.

\bibitem{Wang1}
J.~Wang, Z.~Liu, Y.~Wu, J.~Yuan, Mining actionlet ensemble for action
  recognition with depth cameras, in: IEEE Conference on Computer Vision and
  Pattern Recognition (CVPR), 2012, pp. 1290--1297.

\bibitem{Wang2}
J.~Wang, Z.~Liu, J.~Chorowski, Z.~Chen, Y.~Wu, Robust 3d action recognition
  with random occupancy patterns, Springer,Computer Vision–ECCV (2012)
  872–885.

\bibitem{Shotton}
J.~Shotton, A.~Fitzgibbon, M.~Cook, T.~Sharp, M.~Finocchio, R.~Moore,
  A.~Kipman, A.~Blake, Real-time human pose recognition in parts from single
  depth images, in: IEEE Conference on Computer Vision and Pattern Recognition
  (CVPR), 2011, pp. 1297--1304, dOI:10.1109/CVPR.2011.5995316.

\bibitem{Yao}
H.~Yao, X.~Jiang, T.~Sun, S.~Wang, 3d human action recognition based on the
  spatial-temporal moving skeleton descriptor, in: IEEE International
  Conference on Multimedia and Expo (ICME), 2017, pp. 937--942,
  dOI:10.1109/ICME.2017.8019498.

\bibitem{Yang-Xiaodong1}
X.~Yang, Y.~Tian, Eigenjoints-based action recognition using
  naïve-bayes-nearest-neighbor, in: IEEE Computer Society Conference on
  Computer Vision and Pattern Recognition Workshops (CVPRW), 2012, pp. 14--19,
  dOI: 10.1109/CVPRW.2012.6239232.

\bibitem{Zanfir}
M.~Zanfir, M.~Leordeanu, C.~Sminchisescu, The moving pose: An efficient 3d
  kinematics descriptor for low-latency action recognition and detection, in:
  IEEE International Conference on Computer Vision (ICCV), 2013, dOI
  10.1109/ICCV.2013.342.

\bibitem{Jalal}
A.~Jalal, Y.-H. Kim, Y.-J. Kim, S.~Kamal, D.~Kim, Robust human activity
  recognition from depth video using spatiotemporal multi-fused features,
  Pattern Recognition 61 (2017) 295--308,
  dOI:https://doi.org/10.1016/j.patcog.2016.08.003.

\bibitem{Vieira}
A.~W. Vieira, E.~R. Nascimento, G.~L. Oliveira, Z.~Liu, M.~F. Campos, Stop:
  Space-time occupancy patterns for 3d action recognition from depth map
  sequences 7441 (2012) 252--259, dOI: 10.1007/978-3-642-33275-3-31.

\bibitem{Ellis}
C.~Ellis, S.~Z. Masood, M.~F. Tappen, J.~J. Laviola~Jr, R.~Sukthankar,
  Exploring the trade-off between accuracy and observational latency in action
  recognition, International Journal of Computer Vision 101 (2013) 420--436.

\bibitem{Lv}
F.~Lv, R.~Nevatia, Recognition and segmentation of 3-d human action using hmm
  and multi-class adaboost, in: Proceedings of the Conference on Computer
  Vision-ECCV, Vol.~5, 2006, pp. 359-- 372, dOI:https://doi.org/10.1007/1174.

\bibitem{Peng}
X.~Peng, C.~Schmid, Multi-region two-stream r-cnn for action detection 9911
  (2016) 744--759.

\bibitem{Shou}
Z.~Shou, D.~Wang, S.-F. Chang, Temporal action localization in untrimmed videos
  via multi-stage cnns, in: Proceedings of the IEEE Conference on Computer
  Vision and Pattern Recognition, 2016, pp. 1049--1058, dOI:
  10.1109/CVPR.2016.119.

\bibitem{Wang-Limin}
L.~Wang, Y.~Xiong, D.~Lin, L.~Van~Gool, Untrimmed nets for weakly supervised
  action recognition and detection, in: Proceedings of the IEEE Conference on
  Computer Vision and Pattern Recognition, 2017, dOI: 10.1109/CVPR.2017.678.

\bibitem{Singh}
B.~Singh, T.~K. Marks, M.~Jones, O.~Tuzel, M.~Shao, A multi-stream
  bi-directional recurrent neural network for fine-grained action detection,
  in: Proceedings of the IEEE Conference on Computer Vision and Pattern
  Recognition, 2016, dOI: 10.1109/CVPR.2016.216.

\bibitem{Dave}
A.~Dave, O.~Russakovsky, D.~Ramanan, Predictive-corrective networks for action
  detection, in: Proceedings of the IEEE Conference on Computer Vision and
  Pattern Recognition, 2017, dOI: 10.1109/CVPR.2017.223.

\bibitem{Ma}
S.~Ma, L.~Sigal, S.~Sclaroff, Learning activity progression in lstms for
  activity detection and early detection, in: Proceedings of the IEEE
  Conference on Computer Vision and Pattern Recognition, 2016, dOI:
  10.1109/CVPR.2016.214.

\bibitem{Li-Yanghao}
Y.~Li, C.~Lan, J.~Xing, W.~Zeng, C.~Yuan, J.~Liu, Online human action detection
  using joint classification- regression recurrent neural networks 9911 (2016)
  203--220.

\bibitem{Parisi1}
G.~I. Parisi, C.~Weber, S.~Wermter, Self-organizing neural integration of
  pose-motion features for human action recognition, Frontiers in Neurorobotics
  9, dOI:10.3389/fnbot.2015.00003 (2015).

\bibitem{Parisi2}
G.~I. Parisi, J.~Tani, C.~Weber, S.~Wermter, Lifelong learning of human actions
  with deep neural network self-organization, Neural Networks 96,
  dOI:https://doi.org/10.1016/j.neunet.2017.09.001 (2017).

\bibitem{Wang-Pichao}
P.~Wang, W.~Li, Z.~Gao, J.~Zhang, C.~Tang, P.~O. Ogunbona, Action recognition
  from depth maps using deep convolutional neural networks, IEEE Transactions
  on Human-Machine Systems 46 (2015) 498 -- 509, dOI:10.1109/THMS.2015.2504550.

\bibitem{Hou}
Y.~Hou, Z.~Li, P.~Wang, W.~Li, Skeleton optical spectra based action
  recognition using convolutional neural networks, IEEE Transactions on
  Circuits and Systems for Video Technology (2016)
  1--5DOI:10.1109/TCSVT.2016.2628339.

\bibitem{Ijjina}
E.~P. Ijjina, K.~Mohan~C, Classification of human actions using pose-based
  features and stacked auto encoder, Pattern Recognition Letters 83 (2016)
  268--277, dOI:https://doi.org/10.1016/j.patrec.2016.03.021.

\bibitem{Weinzaepfel}
P.~Weinzaepfel, Z.~Harchaoui, C.~Schmid, Learning to track for spatio-temporal
  action localization, in: Proceedings of the IEEE International Conference on
  Computer Vision, 2015, pp. 3164--3172, dOI:10.1109/ICCV.2015.362.

\bibitem{Li-Ruihao}
R.~Li, D.~Gu, Q.~Liu, Z.~Long, H.~Hu, Semantic scene mapping with
  spatio-temporal deep neural network for robotic applications, Cognitive
  ComputationDOI:https://doi.org/10.1007/s12559-017-9526-9 (2017).

\bibitem{Ding}
S.~Ding, X.~Xi, Z.~Liu, H.~Qiao, B.~Zhang, A novel manifold regularized online
  semi-supervised learning model, Cognitive
  ComputationDOI:https://doi.org/10.1007/s12559-017-9489-x (2017).

\bibitem{MSR}
Y.~W. Wan, Msr action recognition datasets and codes,
  \url{http://research.microsoft.com/en-us/um/people/zliu/actionrecorsrc/}
  (accessed 2015).

\bibitem{balkenius}
C.~Balkenius, J.~Mor\'en, B.~Johansson, M.~Johnsson, Ikaros: Building cognitive
  models for robots, Advanced Engineering Informatics 24~(1) (2010) 40--48,
  dOI:https://doi.org/10.1016/j.aei.2009.08.003.

\bibitem{Johnsson3}
M.~Johnsson.
\newblock \href{http://magnusjohnsson.se/}{[link]}.
\newline\urlprefix\url{http://magnusjohnsson.se/}

\bibitem{Gharaee1}
Z.~Gharaee, A.~Fatehi, M.~S. Mirian, M.~N. Ahmadabadi, Attention control
  learning in the decision space using state estimation, International Journal
  of Systems Science (IJSS) 47 (2014) 1659--1674, dOI:
  10.1080/00207721.2014.945982.

\bibitem{Hesslow}
G.~Hesslow, Conscious thought as simulation of behaviour and perception, Trends
  in Cognitive Sciences 6 (2002) 242--247,
  dOI:https://doi.org/10.1016/S1364-6613(02)01913-7.

\bibitem{johnsson1}
M.~Johnsson, C.~Balkenius, G.~Hesslow, Associative self-organizing map, in:
  Proceedings of the International Joint Conference on Computational
  Intelligence (IJCCI), 2009, pp. 363--370.

\end{thebibliography}


\end{document}